\documentclass[10pt,twocolumn,letterpaper]{article}

\usepackage{iccv}
\usepackage{times}
\usepackage{epsfig}
\usepackage{graphicx}
\usepackage{amsmath}
\usepackage{amssymb}

\usepackage{booktabs}
\usepackage{algorithm}
\usepackage{algorithmic}
\usepackage{makecell}
\usepackage{multirow}

\usepackage{colortbl}
\usepackage{xcolor}
\usepackage{color}

\usepackage{tikz}
\usepackage{diagbox}

\usepackage[accsupp]{axessibility}  % Improves PDF readability for those with disabilities.

\usepackage[pagebackref=true,breaklinks=true,colorlinks,bookmarks=false]{hyperref}

\iccvfinalcopy % *** Uncomment this line for the final submission

\def\iccvPaperID{} % *** Enter the ICCV Paper ID here

\ificcvfinal\fi

\begin{document}

%%%%%%%%% TITLE
\title{Hiding Visual Information via Obfuscating Adversarial Perturbations}

\author{Zhigang Su$^{1,\ast}$\;\;\;\;\;
Dawei Zhou$^{1,}$\thanks{Equal contributions.\;\;$\dagger$ Corresponding author.}\;\;\;\;\;
Nannan Wang$^{1,\dagger}$\;\;\;\;\;
Decheng Liu$^{1,\dagger}$\\
Zhen Wang$^2$\;\;\;\;\;
Xinbo Gao$^3$\\
$^1$Xidian University, $^2$Zhejiang Lab\\
$^3$Chongqing University of Posts and Telecommunications\\
{\tt\small \{zgsu, dwzhou\}.xidian@gmail.com, \{nnwang, dchliu\}@xidian.edu.cn}\\ 
{\tt\small wangzhen@zhejianglab.com, gaoxb@cqupt.edu.cn}
}
% \author{Zhigang Su\\
% Xidian University\\
% {\tt\small zgsu.xidian@gmail.com}
% \and
% Dawei Zhou\\
% Xidian University\\
% {\tt\small dwzhou.xidian@gmail.com}
% \and
% Decheng Liu\\
% Xidian University\\
% {\tt\small dchliu@xidian.edu.cn}
% \and
% Nannan Wang\\
% Xidian University\\
% {\tt\small nnwang@xidian.edu.cn}
% \and
% Zhen Wang\\
% Zhejiang Lab\\
% {\tt\small wangzhen@zhejianglab.com}
% \and
% Xinbo Gao\\
% Chongqing University of\\Posts and Telecommunications\\
% {\tt \small gaoxb@cqupt.edu.cn}
% }

\maketitle
\ificcvfinal\thispagestyle{empty}\fi

%%%%%%%%% ABSTRACT
\begin{abstract}
Growing leakage and misuse of visual information raise security and privacy concerns, which promotes the development of information protection.
Existing adversarial perturbations-based methods mainly focus on the de-identification against deep learning models. %without affecting visual information.
However, the inherent visual information of the data has not been well protected.
In this work, inspired by the Type-I adversarial attack, we propose an Adversarial Visual Information Hiding (AVIH) method to protect the visual privacy of data.
Specifically, the method generates \textbf{obfuscating adversarial perturbations} to obscure the visual information of the data. Meanwhile, it \textbf{maintains the hidden objectives to be correctly predicted by models}. In addition, our method does not modify the parameters of the applied model, which makes it flexible for different scenarios. Experimental results on the recognition and classification tasks demonstrate that the proposed method can effectively hide visual information and hardly affect the performances of models. The code is available at \href{https://github.com/suzhigangssz/AVIH}{https://github.com/suzhigangssz/AVIH}.
\end{abstract}

%%%%%%%%% BODY TEXT
\section{Introduction}
\label{sec:intro}

Deep neural networks (DNNs) have been widely applied in the computer vision \cite{lecun1998gradient,he2016deep,2017Mask}. However, the increasing leakage and misuse of visual information has raised serious concerns \cite{wang2022deepfake,kaur2020comprehensive}, especially in fields such as face \cite{li2022towards, li2023unconstrained, zhou2021towards} and medicine \cite{8357580, alom2019recurrent, hu2021advanced}.
A representative case is the security issue of data stored in the cloud environment \cite{ref3,ref4,ref5}. Due to potential vulnerabilities in cyberspace \cite{ref1}, uploaded private images can be easily stolen and used maliciously \cite{ref2}.
Therefore, it is meaningful and urgent to explore effective strategies to protect visual information.
%Due to the cloud environments are typically untrustworthy \cite{}, the data in the cloud environment is at greater risk of leakage \cite{}. 
%In this work, we focus on hiding visual information while maintaining the correct predictions of particular DNNs.
%In recent years, deep neural networks (DNNs) have been deeply researched and achieved great achievements. To enable DNN-based artificial intelligence systems to provide extensive and convenient services, more and more DNNs are deployed in cloud environments, bringing efficient access to various users. But for tasks like metric learning tasks \cite{ref18,ref19,ref56,ref55} (e.g., face recognition, person re-identification, and 3D shape retrieval), this requires developers or users to upload their private data to the cloud for storage as gallery sets. However, the cloud environments are typically untrustworthy \cite{ref1}, with problems such as data leakage and identity theft \cite{ref2}. Therefore, how to effectively protect the private information in the cloud has always been a common concern \cite{ref3,ref4,ref5}. In this paper, we mainly focus on the visual information hiding of stored images, which is crucial and urgent in real-world scenarios.

A classic strategy is visual information hiding, which mainly consists of two types \cite{ref7}: Homomorphic Encryption (HE)-based methods \cite{ref11,ref12,ref13}, Perceptual Encryption (PE)-based methods \cite{ref6,ref7,ref8,ref9,ref10} and steganography method \cite{kishore2021fixed,zhang2019steganogan}. Affected by the nonlinear activation functions in DNNs, HE-based methods are difficult to perform well on advanced DNNs \cite{ref7}. Although PE-based methods apply to DNNs, existing methods typically require retraining with data in the encrypted domain to guarantee accuracy on encrypted data \cite{ref6,ref7}. This affect the performance of service model on raw data and cause additional resource consumption (especially for large models). The steganography method can hide the visual information of sensitive data into another image, but it does not guarantee that the service model can correctly recognize the protected image.

%-------------------------------------------------------------------------
\begin{figure}[t]
\begin{center}
\includegraphics[width=0.98\linewidth]{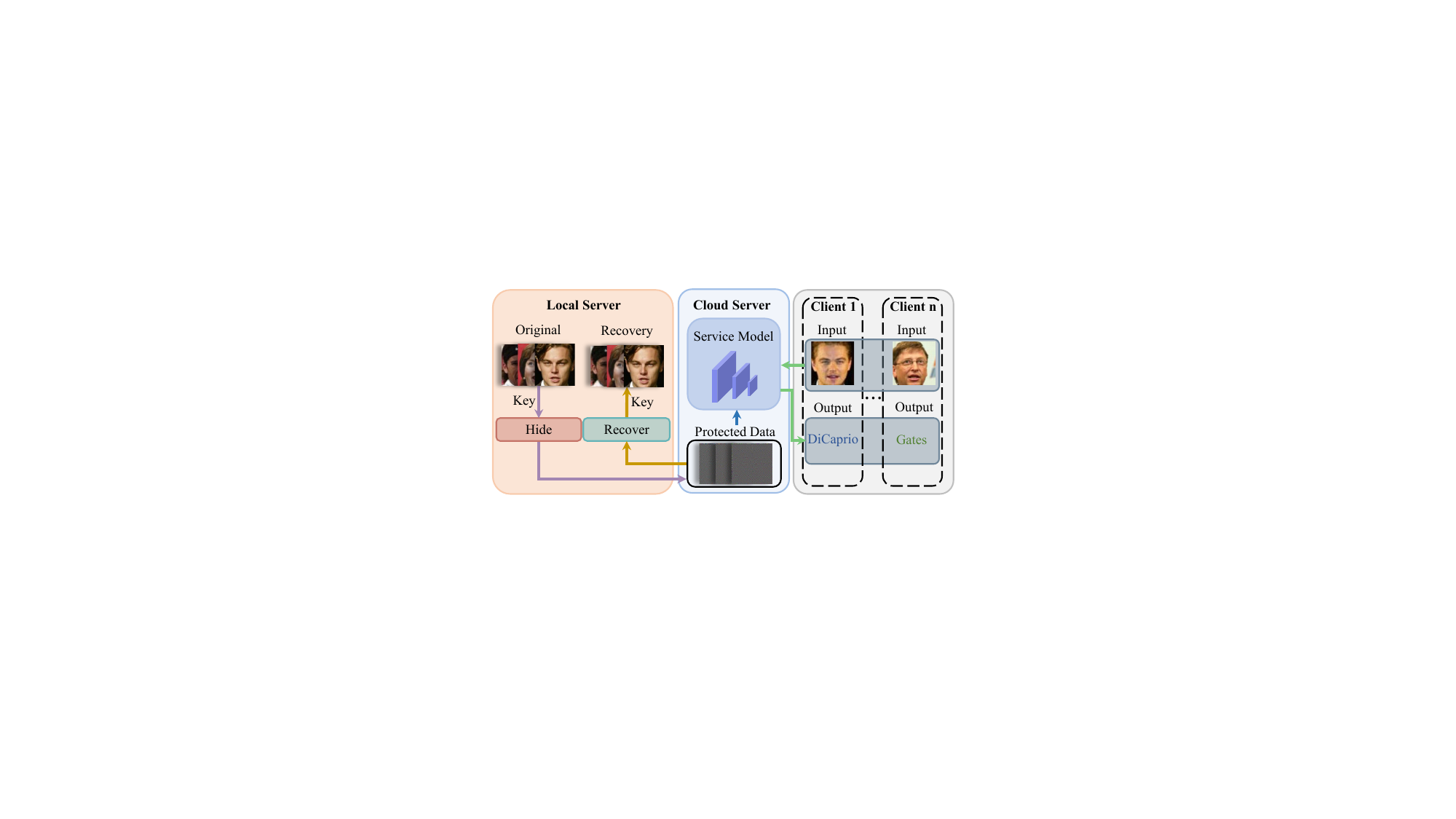}
\end{center}
\caption{Illustration of visual information hiding for face recognition systems in cloud environments. The gallery set is protected and provided to the DNN in the cloud. Protected image has quite different visual information from the original image, but it is still correctly identified. The protected gallery set can be recovered by a key model by the owner.}
\label{fig1}
\end{figure}
%-------------------------------------------------------------------------

To alleviate these negative effects, we expect to hide the visual information without making any modifications to service model. Namely, we hide sensitive visual information only by varying the input image. Previous researches have made some explorations in this regard. The transformation network-based methods \cite{ref10}, which try to protect the original image via a transformation function parameterized by a neural network, share the same philosophy. However, the method cannot easily recover the original image from the protected image for other purposes. They may suffer from adversarial vulnerability \cite{ref14, ref28} because the introduced neural network may be destroyed by adversarial attacks.

In fact, the negative effects of adversarial attacks can be utilized positively to protect privacy. Some works have exploited adversarial attacks for de-identification \cite{ref54,ref16,ref15}. These methods add imperceptible perturbations or non-suspicious patches generated by adversarial attacks to original images, hindering DNNs to extract effective features and recognize identities, thus protecting the identity privacy in the image. However, in this work, we focus on visual information hiding, which means that the protected image is entirely different from the original image visually but can still be correctly predicted by DNNs (see Figure.~\ref{fig1}). Fortunately, we observe a special type of adversarial attack (called Type-I attack) \cite{ref17} quite different from the type used in previous methods. This type of attack guides DNNs to make consistent predictions on two distinct samples.

Inspired by the Type-I adversarial attack, in this paper, we propose an \textit{Adversarial Visual Information Hiding} (AVIH) method. The proposed method hides the visual information in images while preserving their functional (\eg, recognition and classification) features for service model. It can recover original images from protected images for their owners. 
Specifically, we reduce the visual correlation between the protected and original image while minimizing their distance in the feature space of service model, to generate the protected image. 
Meanwhile, we exploit a generative model pre-trained in a private training setting as the key model, then optimize the protected image based on it so that the recovered image is similar to the original image. 
%\blue{Also in the process of generating encrypted images, we introduce a pre-trained generative model as a key.} 
Furthermore, to break through the tough trade-off between the capability of privacy protection and the quality of restored image, we design the variance consistency loss to enhance privacy protection without compromising image recovery (see Section.~\ref{section3.3}). 
Note that the protected image generated by our method can only be accurately recovered by the own key model, other models (even if the model architecture is the same) are difficult to recover well (see Section.~\ref{section4.3}).
%----------------------------------
%The AVH method can effectively protect the visual information of images in the cloud environment. Taking the face recognition system in the cloud environment as an example, the face recognition system requires the manager of the face database to upload the face images as a gallery set to compare with the face images uploaded by the user for face recognition. However, because the cloud environment is often untrustworthy, the face database uploaded to the cloud environment has a great risk of leakage. If the manager only uploads the extracted features to the cloud environment, it is not conducive to the improvement and maintenance of the face recognition system, and the uploaded data cannot be used for other work. Therefore, the owner of the face data can hide the visual information of the data, and then upload the hidden image to the cloud environment. These uploaded data can be accurately extracted by the face recognition model in the cloud environment for normal face recognition tasks. When needed, these information-hiding data can be recovered by a specific key model for other applications. The specific process is shown in Figure.~\ref{fig1}.

AVIH can significantly improve the security and flexibility of image storage, which is extremely obvious in the protection of gallery sets for metric learning. Take the cloud-based face recognition system as an example. According to the service face recognition model, the face database manager can generate protected images locally or in the cloud and save the key model locally. The protected image contains no visual information and can be used by the service model to extract features correctly. Moreover, these protected images cannot be recognized by other models.
%The manager of the face database can encrypt the data locally (or directly in the cloud) based on the target face recognition model, so that the encrypted image does not contain visual information and can be correctly extracted features by the target model. 
Then, these protected images can be stored in the gallery set in the cloud for normal face recognition tasks. These protected images can be recovered using the key model when needed (such as maintaining the dataset or using face images for other tasks, etc.). The process is shown in Figure.~\ref{fig1}.
%----------------------------------
%We demonstrate the effectiveness of our proposed method with comprehensive experiments on visual information hiding of gallery set images for cloud-based face recognition systems. Experimental results on multiple datasets show that the proposed method is effective. In addition, to prove the effectiveness of our proposed loss, we conduct ablation study about it to further present the advantages of our proposed method. To demonstrate that our framework can be applied to a wide range of deep learning tasks, we also implement our method on classification tasks and achieve better results compared to several existing methods.

Taking visual information hiding of gallery set images for cloud-deployed face recognition systems as the basic task, we conduct a comprehensive evaluation of our proposed method in terms of both effectiveness and security. In order to compare with existing methods suitable for information hiding tasks, we extend our method to classification tasks. Experimental results on multiple service models and datasets show that the proposed method is effective. In addition, to prove the effectiveness of our proposed loss, we conduct an ablation study about it to further present the advantages of our proposed method.

%---------------------------------
%To demonstrate the effectiveness of the proposed method, we conduct comprehensive experiments on face recognition tasks and classification tasks. For the face recognition task, we deeply explore our method in terms of both effectiveness and safety. For classification tasks, we compare with other existing methods that can perform visual information hiding.
%To demonstrate the effectiveness of the proposed method, we conduct comprehensive experiments on face recognition tasks and classification tasks. For the face recognition, we encrypt the gallery set on the cloud. Clients can upload their private data for recognition. For the classification task, we encrypt the data transmitted to the classification model. Experimental results on multiple datasets show that the proposed method is effective. In addition, to prove the effectiveness of our proposed loss, we conduct ablation study about it to further present the advantages of our proposed method.

Our main contributions are as follows:
\begin{itemize}
\item{Inspired by Type-I adversarial attacks, we propose a visual information hiding method AVIH. To alleviate the difficult trade-off between capability of information hiding and quality of recovered image, we design a variance consistency loss.}
%to better hide visual information.
\item{Our proposed method has following properties: 1) The visual information in the image is clearly obfuscated. 2) Our method does not require retrain. 3) The protected image can be recovered by the own key model, but external models are difficult to recover.}
\item{We validate the effectiveness of the proposed method on the face recognition task and the classification task. In addition, we conduct qualitative and quantitative ablation studies to show efficiency of the proposed loss.}
\end{itemize}

%-------------------------------------------------------------------------
\begin{figure*}
  \begin{center}
  \includegraphics[width=0.98\textwidth]{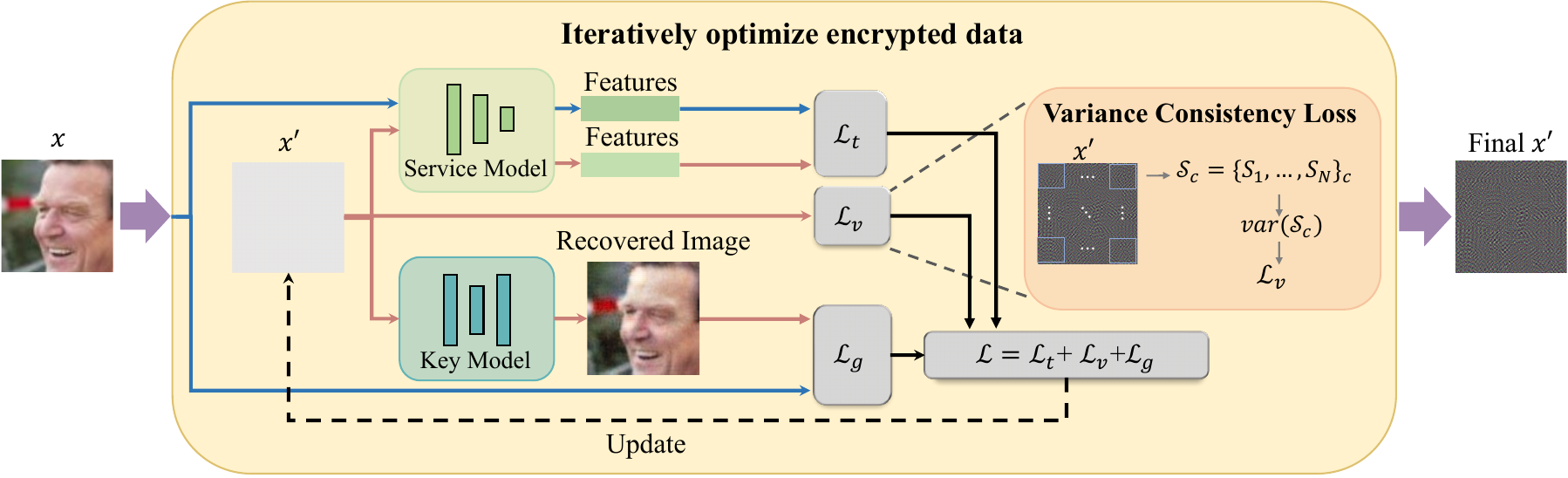}
  \end{center}
  \caption{Overview of adversarial attack-based visual information hiding (AVIH) method. Taking the face recognition as an example, given a service model and a pre-trained key model, we protect the original image $x$ and obtain the protected image $x^{\prime}$.}
   \label{fig2}
\end{figure*}

\section{Related Work}
\subsection{Visual Information Hiding}
The visual information hiding of images is from the perspective of human vision. It is most directly manifested by the protected images being visually unrecognizable. Since Homomorphic Encryption (HE) only supports additive and multiplicative operations, HE are not suitable for nonlinear computations. Most DNNs contain a large number of nonlinear computations, so HE-based methods are hardly applicable to state-of-the-art DNNs \cite{ref7}. The steganography methods hide sensitive information in a cover image and require as few changes to the cover image as possible. It does not need to guarantee that the service model can perform normal predictions on hidden information, which is very different from our work. Therefore, these two types of methods are not discussed in this paper. For the methods based on Perceptual Encryption (PE), some works \cite{ref6,ref7,ref8,ref9} focus on finding an encrypted domain and training the model directly using the encrypted images. However, this has a significant impact on the accuracy of the model. To improve the accuracy of the classification model for protected images, Ito \etal \cite{ref10} trained a transformation network to keep the classifier correctly classified while hiding visual information. However, the weakness of this method is that the protected image cannot be recovered.

Unlike the current work used in classification and segmentation tasks, we mainly focus on metric learning tasks represented by face recognition, which are more prone to privacy leakage. To compare with existing methods that can achieve visual information hiding, we also extend the AVIH method to classification tasks. Our proposed method can also compensate for the drawbacks of the above methods. It generates protected images for a specific model that already exists. It takes advantage of the vulnerability in the model itself to provide strong privacy protection to the image. Since our method does not modify the service model, it does not affect the accuracy on the original image.

\subsection{Adversarial Attack}
DNNs are vulnerable to some adversarial examples \cite{ref14,ref28}. There are many adversarial attack methods \cite{ref14,ref23,ref26} to find adversarial examples efficiently. Tang \etal \cite{ref17} divided the adversarial attacks into adversarial attack Type-I and adversarial attack Type-II based on the statistical Type I and Type II error. We take an optimization perspective on adversarial attacks. Then the Type-II attack maximizes the difference in the model output while ensuring a slight difference with the original input samples. Mathematically, it can be formulated as follows:
\begin{equation}
  \label{ex1}
  \begin{array}{*{20}{c}}
    {\mathop {\max }\limits_{x'} f(x)-f({x}')}&{s.t.}&{\left\| {x' - x} \right\| < \epsilon}
    \end{array},    
  \end{equation}
where $x$ is the original sample, ${x'}$ is the adversarial sample, and $f( \cdot )$ is the model which is attacked. The Type-I attack, in contrast to the Type-II attack, looks for an input sample that differs the most from the original input sample but makes the model output as same as possible. Mathematically, it can be formulated as follows:
\begin{equation}
  \label{ex2}
  \begin{array}{*{20}{c}}
    {\mathop {\max }\limits_{x'} \left\| {x' - x} \right\|}&{s.t.}&{f(x) = f(x')}
    \end{array}.    
  \end{equation}
The Type-I attack on classification and generative models was implemented in the work of Tang \etal \cite{ref17}. Then, Sun \etal \cite{ref29} implemented a Type-I attack against variational autoencoder (VAE) \cite{ref30}. In this work, we implement Type-I attacks on face recognition and classification tasks. Inspired by these attacks, we propose the AVIH method.

\section{Methodology}
The objective of our proposed method is to learn a protected image ${{x}^{\prime}}$ which satisfies: 1) ${{x}^{\prime}}$ is completely different from the original image ${x}$, 2) for the service model the output ${{f}_{s}}({{x}^{\prime}})$ is the same as the ${{f}_{s}}(x)$, and 3) ${{x}^{\prime}}$ can be recovered as the ${x}$ by the ${key}$ model. Mathematically,
\begin{equation}
  \label{ex3}
    \begin{gathered}
\mathrm{From} \quad  x \in {\mathcal X} \quad \ \mathrm{Generate}\quad x^{\prime}= \mathcal A(x)\\
	s.t. \; \; \; \; \; \; \; \; \; \; \; \;  \quad\ \left\| {{x^{\prime}} - x} \right\| > \epsilon \; \; \; \; \; \; \; \; \; \; \; \; \; \; \; \; \; \; \\
	\; \; \;{{f_s}({x^{\prime}}) = {f_s}(x)}, \; {key({x^{\prime}}) = x}.
    \end{gathered}
  \end{equation}

The pipeline of our method is shown in Figure.~\ref{fig2}.

\subsection{Image Visual Information Hiding}
Exploiting the adversarial vulnerability of DNNs, we can perform Type-I attack on the service model to get images that are visually completely different from the original image but with extremely similar features.
Suppose that ${{d( \cdot )}}$ measures the difference between the outputs of the model. For different tasks, it behaves as different functions. Then for a particular service model ${{f}_{s}}$, the difference between the output of the adversarial sample and the original sample is 
\begin{equation}
  \label{ex4}
  {{\mathcal{L}}_{t}}({{x}^{\prime}},{x})={d({f}_{s}(x), {f}_{s}({x}^{\prime}))}.  \end{equation}
We definite ${{\mathcal{L}}_{d}}$ as
\begin{equation}
  \label{ex5}
  {{\mathcal{L}}_{d}}({{x}^{\prime}},{x})={{\left\| {x}-{{x}^{\prime}} \right\|}_{2}^{2}} ,
  \end{equation}
then the loss of the Type-I attack on the service model is formulated as
\begin{equation}
  \label{ex6}
    {{\mathcal L}_r}({x^{{\prime}}},{x}) = {{\mathcal{L}}_{t}}({{x}^{\prime}},{x}) - \lambda \cdot {{\mathcal L}_d}({{{x}^{\prime}},{x}}),
  \end{equation}
 where $\lambda$ is a positive hyperparameter which balances image level differences and output level differences. Then with a certain number of iterations $K$, we can optimize $\mathcal{L}_r$ by the following operations to get the final adversarial sample.
\begin{equation}
  \label{ex7}
  {{g}_{k+1}}=\alpha \cdot {{g}_{k}}+\frac{\nabla \mathcal{L}({{x}_{k}^{\prime}},{x})}{{{\left\| \nabla \mathcal{L}({{x}_{k}^{\prime},{x}}) \right\|}^{2}_{2}}},
\end{equation}
\begin{equation}
  \label{ex8}
  x_{k+1}^{\prime}=x_{k}^{\prime}-\beta \cdot {{g}_{k+1}},
\end{equation}
where
\begin{equation}
  \label{ex16}
  {{g}_{0}}=\frac{\nabla \mathcal{L}({{x}_{0}^{\prime}},{x})}{{{\left\| \nabla \mathcal{L}({{x}_{0}^{\prime},{x}}) \right\|}^{2}_{2}}}.
\end{equation}
At the first iteration, ${{x}_{0}^{\prime}}$ can be randomly initialized, or it can be made ${{x}_{0}^{\prime}}={x}$. The former provides an easier way to get an adversarial example with more significant visual differences. The latter provides a faster way to get an adversarial example that meets the criteria.

\subsection{Image Recovery}
\label{section3.2}
The protected image obtained by Equation.~\ref{ex6} can effectively hide visual information and keep the function of the sensitive information for the service model. However, the protected image is not recoverable. To achieve the goal of Equation.~\ref{ex3}, we attach a recovery module that can recover the generated information hidden image. We refer to the generated images with perturbations as protected images. This perturbation has the function of protecting the visual information of the image.
%Since the process of generating the information hidden and recovering the information hidden image is equivalent to a process of encryption and decryption, we refer to the generated information hidden image as an encrypted image and the generation model that can recover the encrypted image as the key model.

To get the protected images, we first train the generative model $G$, which can generate the same images as the input. 
%In this work, we use a generative model based on the pix2pix framework \cite{ref32}. 
Then we perform Type-I attacks on both the service model and the generative model. We define ${{\mathcal{L}}_{g}}$ as follows:
\begin{equation}
 \label{ex9}
{{\mathcal L}_g}({x^{\prime}},{x}) = {\left\| {{x} - G({x^{\prime}})} \right\|^2_2}.
\end{equation}
The loss ${\mathcal L}_g$ can help to keep the protected images recoverable by the key model we have chosen. Thus, the loss becomes as follows:
\begin{equation}
\label{ex10}
{{\mathcal L}_e}({x^{\prime}},{x}) = {{\mathcal L}_r}({x^{\prime}},{x}) + \mu \cdot {{\mathcal L}_g}({x^{\prime}},{x}),
\end{equation}
where $\mu $ is a hyperparameter that balances the protection quality and restore quality. With Equation.~\ref{ex7} and Equation.~\ref{ex8}, we can optimize ${{\mathcal L}_e}$ to obtain the protected image ${{x}^{\prime}}$,which can satisfy the objective of Equation.~\ref{ex3}. 

\subsection{Variance Consistency Loss}
\label{section3.3}
The protected image obtained by the objective function of Equation.~\ref{ex6} satisfies the requirements of Equation.~\ref{ex3}, but its protection quality is not high. The obtained protected images have the problem of the difficult trade-off between protection quality and recovery quality. That is, if we want to obtain an image that is difficult to be cracked successfully, the quality of the image recovered by the key model will be poor. Another problem is that the obtained protected image, although differing greatly from the original in color, has obvious visual information that the original image has, which negatively impacts visual information protection. To solve the above problem, we propose a variance consistency loss. It improves the quality of protection by limiting the differences between each part of the image to make the protected image visually more confusing.

\begin{algorithm}[t]
\caption{Adversarial visual information hiding
method}
\label{alg:algorithm}
%\begin{small}
\textbf{Input}: Service model ${f}_{s}$; key model $G$; original image $x$; number of iterations $K$; gradient $g$; number of times the loss has increased $num$; maximum number of times the loss increase $maxnum$.\\ %\blue{explain g and num} 
\textbf{Output}: Protected images ${x}^{\prime}$.

\begin{algorithmic}[1] %[1] enables line numbers
\STATE ${g}_{0} = 0$; $num = 0$;
\STATE Random initialization ${{x}_{0}^{\prime}}$;
\FOR{$k = 0$ to $K$}
\STATE Compute ${\mathcal L}_{AVIH}({x^{\prime}_{k}},{x})$ via Equation.~\ref{deqn_ex13};
\STATE ${{g}_{k+1}}=\alpha \cdot {{g}_{k}}+\frac{\nabla \mathcal{L}_{AVIH}({{x}_{k}^{\prime}},{x})}{{{\left\| \nabla \mathcal{L}_{AVIH}({{x}_{k}^{\prime},{x}}) \right\|}^{2}_{2}}}$ (Equation.~\ref{ex7});
\STATE $x_{k+1}^{\prime}=x_{k}^{\prime}-\beta \cdot {{g}_{k+1}}$ (Equation.~\ref{ex8});
\IF {${\mathcal L}_{AVIH}({x^{\prime}_{k-1}},{x}) < {\mathcal L}_{AVIH}({x^{\prime}}_{k},{x})$}
\STATE $num = num + 1$;
\ENDIF
\IF{$num > maxnum$} 
\STATE $\beta = 0.85 \cdot \beta$;
\STATE $num = 0$;
\ENDIF
%\STATE ${\mathcal L}_{AVIH}({x^{\prime}},{x})^{former} = {\mathcal L}_{AVIH}({x^{\prime}},{x})$
\ENDFOR
\end{algorithmic}
%\end{small}
\end{algorithm}

For the input image, we divide the image in each channel (R, G, B) into $N$ blocks respectively, \ie, ${\left\{ {{b_{1}},{b_{2}}, \ldots ,{b_{N}}} \right\}_c}$, where $c \in \{R,G,B\}$, ${b_{n}} \in {{\mathbb{R}}^{h \times w}}$ and $h, w$ denote the height and width of the block. The pixels of the blocks are allowed to have overlapping parts between them. Let $p_{i,j}^{{b}_{n}}\in [0,1]$ denote the normalized pixel value at $(i,j)$ in block ${b}_{n}$. Then, we calculate the sum of each block:
\begin{equation}
  \label{ex11}
  {{S}_{n}}=\sum\limits_{i=1}^{h}{\sum\limits_{j=1}^{w}{p_{i,j}^{b_{n}}}} \text{.}
\end{equation}
We use $\mathcal{S}$ denote the set of sum of the blocks for each channel, \ie, ${\mathcal{S}_c}={{\left\{ {{S}_{1}},{{S}_{2}},\ldots ,{{S}_{N}} \right\}}_{c}}$. In our practice, we convolve the image with a convolution kernel of size $h\times w$ to obtain blocks. Then, we calculate the variance of ${\mathcal{S}_c}$ and obtain $\sigma _{c}^{2} = \mathop{\rm var}(\mathcal{S}_{c})$. Finally, we define the variance consistency loss as:
\begin{equation}
  \label{ex12}
  {{\mathcal{L}}_{v}}({{x}^{\prime}})=\sigma _{R}^{2}+\sigma _{G}^{2}+\sigma _{B}^{2},
\end{equation}
where $\sigma _{R}$, $\sigma _{G}$ and $\sigma _{B}$ denote the variances for R, G, B channels, respectively. By minimizing ${{\mathcal{L}}_{v}}$, we can get a protected image with a more uniform distribution of pixel values. Proof-of-concept experiments (see Section.~\ref{section4.4}) show that ${{\mathcal{L}}_{v}}$ can eliminate the visual semantics in the protected images which are similar to the original images, and can help obtain protected images with high quality of protection and recovery. More details on the variance consistency loss can be found in Supplementary Material D.

Based on the variance consistency loss, the loss function in Equation.~\ref{ex6} is modified as: 
\begin{equation}
  \label{ex17}
    {{\mathcal L}_r}({x^{{\prime}}},{x}) = {{\mathcal{L}}_{t}}({{x}^{\prime}},{x}) + \lambda \cdot {{\mathcal L}_v}({{{x}^{\prime}}}),
  \end{equation}
and the Equation.~\ref{ex10} is reformulated as: 
\begin{equation}
\label{deqn_ex13}
\mathcal{L}_{AVIH}({x^{\prime}},{x}) = {\mathcal L}_{t}({x^{\prime}},{x}) + \lambda \cdot {{\mathcal L}_{v}}({x^{\prime}}) + \mu \cdot {{\mathcal L}_{g}}({x^{\prime}},{x}),
\end{equation}
where $\mu$ are hyperparameters. This is our proposed AVIH framework for visual information hiding. The algorithm is summarized in Algorithm.~\ref{alg:algorithm}.

%----------------------------------------------------

%----------------------------------------------------------

\subsection{AVIH Method for Specific Tasks}
%Our proposed AVH method is mainly used for privacy-preserving face recognition tasks, and for other tasks in deep learning, our AVH framework also performs well.
Our method can provide effective protection for the stored image and can be applied to a wide range of tasks. In this work, we take face recognition and classification as examples to illustrate the ability of our method.

For the face recognition task, we want the features extracted by the service model ${f}_{s}$ from the protected and original images to be as same as possible. Thus, we modify ${\mathcal L}_{t}$ as follows:
\begin{equation}
\label{ex15}
{{\mathcal{L}}_{t}}({{x}^{\prime}},{x})={{\left\| {{f}_{s}}({x})-{{f}_{s}}({{x}^{\prime}}) \right\|}^{2}_{2}}.%\blue{MSE?}
\end{equation}

For the classification task, compared with directly minimizing the mean square error (MSE)  of ${{f}_{s}}({x})$ and ${{f}_{s}}({{x}^{\prime}})$, we find that maximizing ${{f}_{s}^c}({{x}^{\prime}})$ where $c$ is the prediction class of the service model on the original image can more effectively reduce the impact of visual information hiding on the accuracy.
We thus first obtain the logit output of the original image ${{f}_{s}}({x})$ and convert it to one-hot format $\delta(f_s(x))$. Then, we reify $\mathcal{L}_{t}$ as follows:
\begin{equation}
\label{ex14}
{{\mathcal L}_{\rm{t}}}({{x}^{\prime}},{x}) = -\delta(f_s(x)) \cdot f_s({x^{\prime}}).
\end{equation}
%-------------------------------------------------------------------------
\section{Experiments}
In this section, we first verify the effectiveness of the AVIH method for face recognition tasks. Then, taking the face recognition system as an example, the security of the AVIH method and the effectiveness of the variance consistency loss are explored. Finally, we verify the effectiveness of the AVIH method for classification tasks and compare it with other methods.
\subsection{Experimental Settings}

\begin{figure*}[t]
\begin{center}
\includegraphics[width=0.98\textwidth]{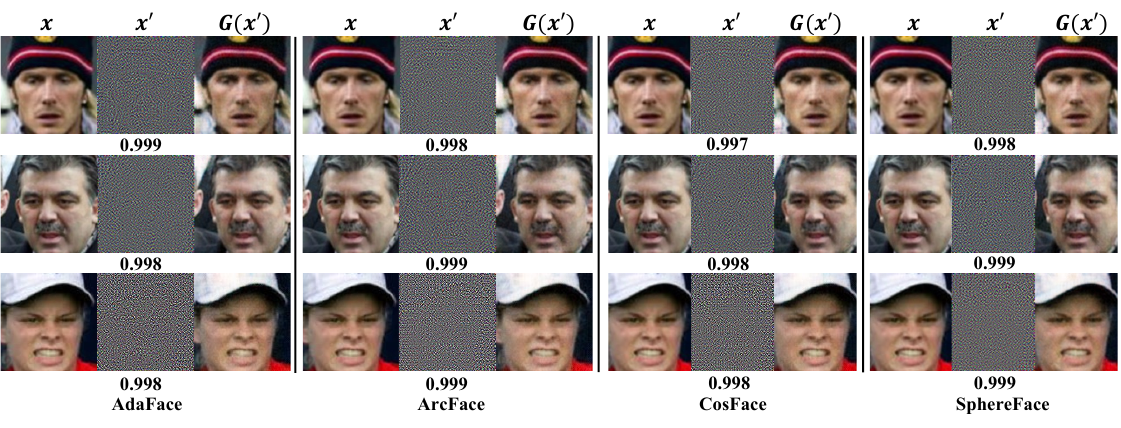}
\end{center}
\caption{Adversarial Visual Information Hiding (AVIH) method protect face images for different service models. We mark below each protected image the cosine similarity between its feature vector and the original image's feature vector.}
\label{fig3}
\end{figure*}

\textbf{Dataset and service models.} Our experiments were mainly evaluated based on the \textit{Labeled Face in the Wild (LFW)} \cite{ref41} dataset. We choose four face recognition models ArcFace \cite{ref18}, CosFace \cite{ref19}, SphereFace \cite{ref20} used in \cite{ref15}, and AdaFace \cite{ref53} to fully evaluate the performance of our method. Among them, the input size of ArcFace and AdaFace \cite{ref53} is $112\times 112$, and the input size of CosFace and SphereFace is $112\times 96$. Therefore, we first use the MTCNN \cite{ref44} to align and crop the face images to the corresponding face recognition model's input size. 

\textbf{Key models.} We choose the pix2pix framework \cite{ref32} to train our key model.
%It has the same structure as the generator used in CycleGAN \cite{ref22}. 
%We use the structure of 9 residual blocks to construct it. 
To pre-train the key model, we randomly selected 1,2878 images from the \textit{CelebA} \cite{ref45} dataset as the training set. Then we set the input and output of the model to the same image, set the batch size to 1, and train for 4 epochs. Finally, the trained generator is used directly as the key model. Unless otherwise stated, we use only one key model to protect the entire dataset.

\textbf{Evaluation metrics.} To evaluate the effectiveness of our method more realistically and inspired by the evaluation method in MegaFace \cite{ref42}, we modified the evaluation method of \textit{LFW}. We randomly selected 12 persons from the \textit{LFW} dataset as the probe set. Each contains more than 12 facial images, comprising 355 images. The other 12878 images, we use as the gallery set. In the testing phase, we take one face image of a person in the probe set and put it into the gallery set. Then use the remaining images of this person as the test set. Next, we use the above-divided dataset to test the accuracy of the face recognition model. In this way, we put each person's image in the probe set into the gallery set in turn to measure the average accuracy. This metric can well demonstrate the impact of our method on face recognition models in practical applications.

We use the Structural Similarity Index Measure (SSIM) \cite{ref46} and Learned Perceptual Image Patch Similarity (LPIPS) to quantitatively measure the quality of the protected images. In this work, we use SSIM$_{e}$ / LPIPS$_{e}$ to represent the average of the SSIM / LPIPS values between the protected image ${x^{\prime}}$ and the original image ${x}$, and use SSIM$_{d}$ / LPIPS$_{d}$ to represent the average of the SSIM / LPIPS values between ${x}$ and the recovered image $G({{x}^{\prime}})$. A higher SSIM indicates that the two images are more similar and a lower LPIPS indicates that the two images are more similar.

\subsection{Effectiveness of Face Privacy Protection}

\textbf{Effectiveness and impact on service model.} To the best of our knowledge, our method is the first (PE)-based method for face recognition. There is no closely related method yet to compare the performance of protected images. Therefore, we conduct qualitative and quantitative evaluations of protected images. High visual metric scores and a very slight impact on accuracy demonstrate the effectiveness of our method.
%The high recognition accuracy and visual metric scores indicate the validity of our method. 

We use the AVIH method to protect the original face image ${x}$ in the probe set to test its effectiveness on face visual information hiding. Then, we used the obtained protected image ${x^{\prime}}$ as the input of the key model $G(\cdot )$. Finally get the recovered image $G({{x}^{\prime}})$. The results are shown in Figure.~\ref{fig3}. From Figure.~\ref{fig3}, it is evident that the protected image generated by our method is significantly different from the original image. Since no useful visual information is obtained from the protected image, it has the effect of visual information hiding for the original image. In Table.~\ref{tab:1}, the average SSIM value between the protected and original images for each service model is less than 0.04, while the cosine similarity between their features is higher than 0.99. Therefore, the protected image generated by AVIH can completely replace the original image while hiding the visual information.

\begin{table}[t]
  \caption{Image visual quality metrics and cosine similarity (cos-sim) between the output features of the service model for original and protected images on \textit{LFW}.}
  \label{tab:1}
  \renewcommand\tabcolsep{2.0pt}
  \renewcommand\arraystretch{1.0}
  \begin{center}
  \begin{small}
  \begin{tabular}{lccccc}
  \toprule
  & SSIM$_{e}$ $\downarrow$& LPIPS$_{e}$$\uparrow$ & SSIM$_{d}$ $\uparrow$ & LPIPS$_{d}$$\downarrow$& cos-sim $\uparrow$\\
  \midrule
  AdaFace & 0.028 &0.886& 0.893 &0.144& 0.997\\
  ArcFace & 0.030& 0.888& 0.903&0.128 & 0.994\\
  CosFace & 0.028& 0.891& 0.899 &0.128& 0.998\\
  SphereFace & 0.029&0.888 & 0.906 &0.120& 0.996\\
  \bottomrule
  \end{tabular}
  \end{small}
  \end{center}
\end{table}

\begin{table}[t]
\caption{Accuracy (percentage) of face recognition models for original image and different protected image on \textit{LFW}. The results of AVIH-ONE and AVIH-ALL are expected to be close to the results of original.}
\label{tab:2}
\renewcommand\tabcolsep{6.0pt}
\renewcommand\arraystretch{1.0}
\begin{center}
\begin{small}
\begin{tabular}{lccc}
\toprule 
Models & Original & AVIH-ONE ($\%$) & AVIH-ALL ($\%$)\\
\midrule
AdaFace & 98.6 & 98.6 & 98.6\\
ArcFace & 96.5 & 96.5 & 96.5\\
CosFace & 89.4 & 89.2 & 89.3\\
SphereFace & 80.3 & 80.0 & 80.6\\
\bottomrule 
\end{tabular}
\end{small}
\end{center}
\vskip -0.03in
\end{table}

To evaluate the impact of AVIH method on the accuracy of the face recognition models, we first generate a protected face image in the probe set before putting it into the gallery set. Then put the protected image into the gallery set instead of the original image. We tested the accuracy of the face recognition models in this way. The result AVIH-ONE is shown in Table.~\ref{tab:2}. Compared with the original accuracy of the models, it can be concluded that the AVIH method's impact on the models' accuracy is very slight, which can achieve the same accuracy in the ArcFace model.

Considering the actual situation, all the images stored in the gallery set are protected images. So we replaced all the images in the gallery set with protected images and then retested the accuracy of the models, as AVIH-ALL in Table.~\ref{tab:2}. In this case, our method has a very slight impact on the accuracy of the models and has a beneficial effect on the accuracy of the SphereFace model.

We test the average of SSIM values, as shown in Table.~\ref{tab:1}. Combined with Figure.~\ref{fig3} shows that the recovered image has good quality and can recover most of the visual information of the original image. Since the recovered images are generated by the generative model, a better-trained generative model can achieve better recovery quality. We also explore the time spent in protecting images of AVIH in Supplementary Materials A.
%We also explore the time taken by AVIH to protect the image in Supplementary Material B.

\textbf{Randomness of the dataset for training the key model.} In this work, we use \textit{CelebA}, which is also a face dataset, to train the key model. However, in real scenarios, large-scale face images are often difficult to obtain due to the privacy of each individual involved. So we changed \textit{CelebA} to \textit{COCO} \cite{ref51} containing various objects to test the effectiveness of the new key model. We choose the test set of \textit{COCO} as the train set and train 6 epochs on the key model. The results are shown in Table.~\ref{tab:3}. It can be concluded that the impact of protected images on the model accuracy and the quality of recovered images are slightly different from the key model trained with \textit{CelebA}. Therefore, we can use different types of datasets to train the key model, not just limited to face images.

\begin{table}[t]
  \caption{The performances of our method when the pre-trained model is based on non-face image (\eg, \textit{COCO}). The numbers in parentheses represent the difference from the original accuracy of the model. For example, (-0.2) means the accuracy is 0.2 lower than the original.}
  \label{tab:3}
  \renewcommand\tabcolsep{8.0pt}
  \renewcommand\arraystretch{1.0}
  \begin{center}
  \begin{small}
  \begin{tabular}{lccccc}
  \toprule 
    Model & AVIH-ONE ($\%$)&SSIM$_{e}$ $\downarrow$&SSIM$_{d}$ $\uparrow$\\
  \midrule
  AdaFace & 98.6(-0.0) & 0.026 & 0.891\\
  ArcFace & 96.5(-0.0)&0.027&0.909\\
  CosFace & 89.2(-0.2)&0.031&0.882\\
  SphereFace & 79.9(-0.4)&0.031&0.890\\
  \bottomrule 
  \end{tabular}
  \end{small}
  \end{center}
  \vskip -0.1in
\end{table}

\begin{figure}[t]
 \begin{center}
 \includegraphics[width=0.98\columnwidth]{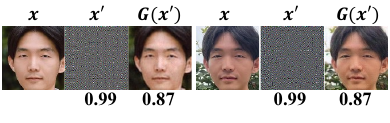}
 \caption{Visual information hiding in real-world scenes. The cosine similarity between recognition features of protected images ($x^{\prime}$) and original images ($x$) is marked below $x^{\prime}$. The SSIM between recovered images ($G(x^{\prime})$) and original images is marked below $G(x^{\prime})$.}
 \label{fig4}
 \end{center}
 \vskip -0.1in
\end{figure}

\textbf{Real world face privacy protection.} We randomly selected real-world face photos taken with phones and then protected them to test the performance of the AVIH method for using real-world scenarios. Part of the results is shown in Figure.~\ref{fig4}. In realistic scenarios, the cosine similarity of the features between the protected image generated by the AVIH method and the original image is still higher than 0.99. The recovered image can still be recovered well.

\textbf{Robustness to noise.} We tested the quality of the recovery after adding Gaussian noise $X_{noise}\sim\mathrm{N}(0,\sigma^2)$ to the protected image, as shown in Table~\ref{tab:A}. It can be concluded that our method is robust to small noise.

\begin{table}[t]
\renewcommand\tabcolsep{8.0pt}
\renewcommand\arraystretch{1.0}
\begin{center}
\caption{Robustness of protected images against noise.}
\vskip 0.1in
\label{tab:A}
\begin{small}
\begin{tabular}{l|ccccc}
\toprule
$\sigma$ & 0 & 3/255 & 5/255 & 8/255 & 10/255\\
\midrule
SSIM$_{d}$ & 0.93 & 0.92 & 0.89 & 0.84 & 0.81\\
\bottomrule 
\end{tabular}
\end{small}
\end{center}
\vskip -0.1in
\end{table}

\begin{figure}[t]
  \begin{center}
  \includegraphics[width=0.98\columnwidth]{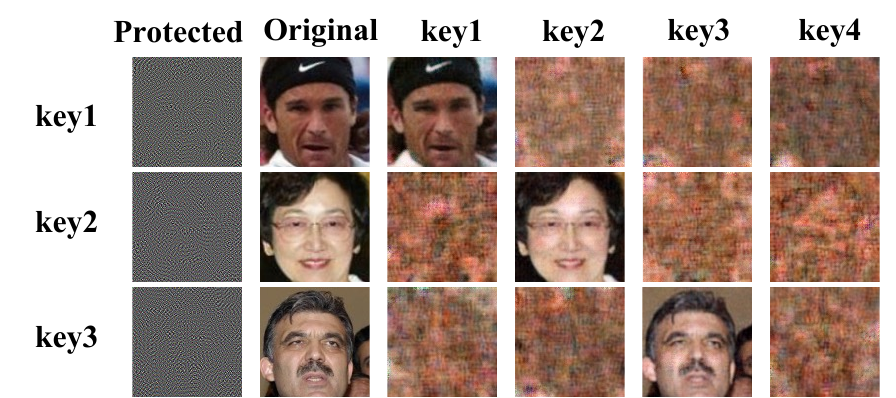}
  \end{center}
  \caption{Partial results of the key model randomness analysis. The left column is the key used for protection, and the top row is the key used for recovery. Every key model is trained with different initialization only. More images of the results are shown in Supplementary Material B.}
  \label{fig5}
\end{figure}

\begin{figure}[t]
  \begin{center}
  \includegraphics[width=0.98\columnwidth]{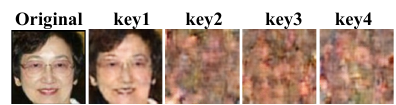}
  \end{center}
  \caption{Crack the protected image using the attack model. The training dataset for the attack model is obtained using key1 protection. An attacker can recover images protected with his own key (key1), but cannot recover images protected with other keys.}
  \label{fig6}
  \vskip -0.1in
\end{figure}

\begin{figure*}[t]
  \begin{center}
  \includegraphics[width=0.99\textwidth]{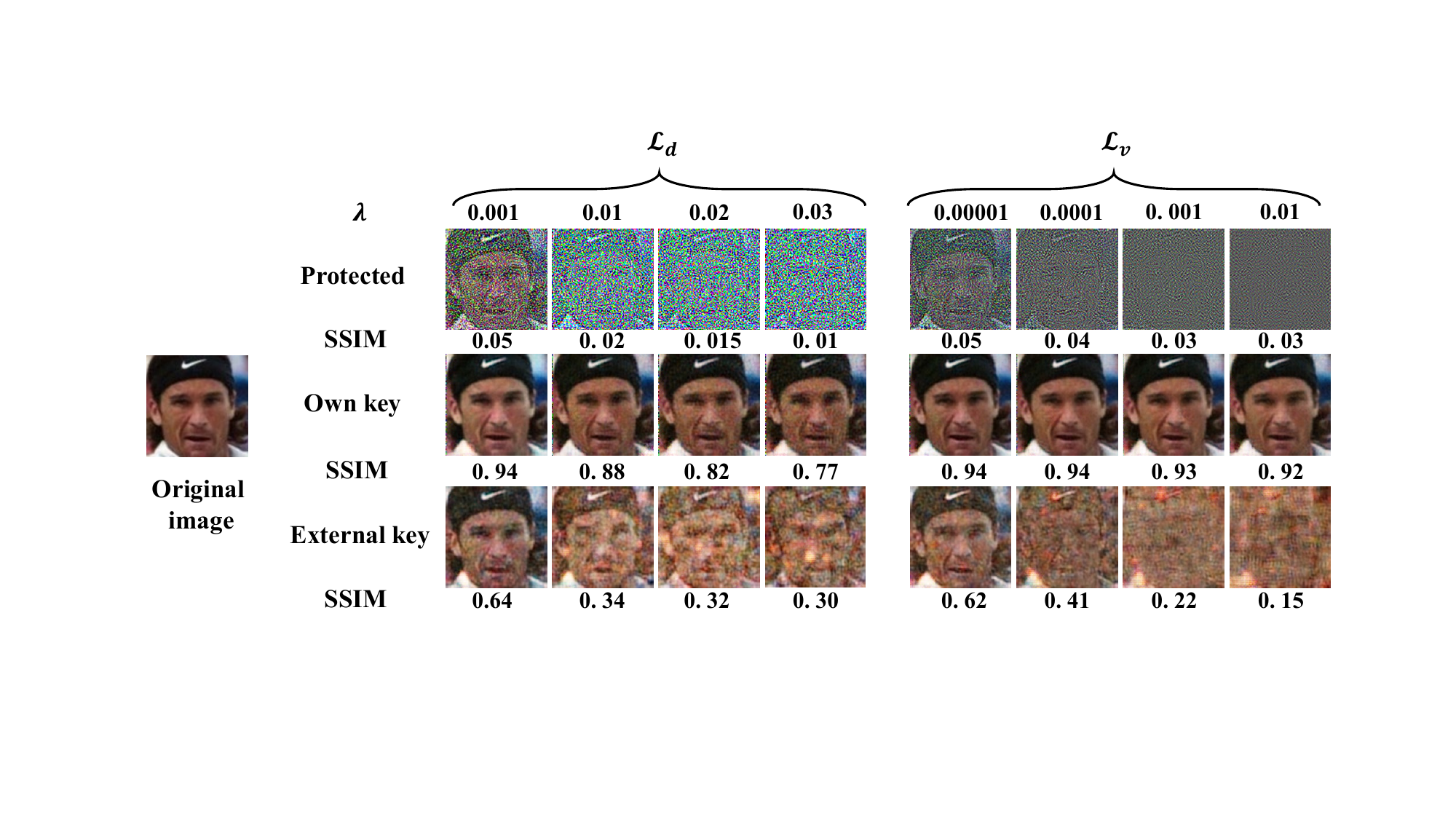}
  \end{center}
  \caption{Results of the ablation study. The SSIM value between the processed image and the original image is shown under each image. The own key represents the key model used for generating the protected image, and the external key represents other key models that are initialized differently during training.}
  \label{fig7}
\end{figure*}

\subsection{Security Analysis}
\label{section4.3}
\textbf{Key model randomness analysis.} To test the randomness of the key model, we use the same strategy as Ding \etal \cite{ref9}. We first trained 16 key models using the same settings but with different initialization values. Different initialization values may result in significantly different key models. Then, we choose one of them in turn as the private key model to participate in the protection of the image, and use the others as the external key model to try to recover the protected image. From the results shown in Figure.~\ref{fig5}, it can be concluded that the protected image obtained by one private key model cannot be recovered by other external key models, demonstrating that our proposed method can greatly improve the security of the face visual information. In addition, it is convenient that we can obtain mutually independent keys by simply changing the initialization without changing factors such as the structure of the key model and the training dataset. We suppose that one reason for this phenomenon is the instability of GAN training, where different initialization values lead to different locally optimal solutions. Another reason could be that our method tends to find the boundary points of the input field corresponding to the output error allowed by the key model. However, the instability of GAN training causes this boundary to change significantly when the initialization value is changed.

\textbf{Responding to a possible attack.} For the key model, even if the key model structure and its training settings are leaked, the protection is still difficult to break if the initialization point of the key training model is not leaked. As for the image, if pairs of original and protected images are leaked in large quantities, then an attacker can directly use these image to train a generative model to directly map the protected domain to the original domain. 
Ito \etal \cite{ref52} exploited this method to evaluate the robustness of protection. So, we protect all the face images in the gallery set using the same key model, producing pairs of protected and original images. Then we use these paired images as a training set, and train a model using the same structure as the attack model. Using such a model, we attempt to recover the protected images in the probe set. The results are shown in Figure.~\ref{fig6}. When the training set images and the protected image to be cracked are protected from the same key model, the attack model can recover the protected image. However, when they are protected with different key models, the trained model cannot crack the protected image. Therefore, we can increase the frequency of changing key models to effectively defend against such attacks. 

The protected images obtained by our method also have good statistical properties. The correlation analysis and histogram analysis are shown in Supplementary Material B. We also verify that the protected image generated by a service face recognition model cannot be recognized by other models with the same function.
%-------------------------------------------------------------------------

\subsection{Ablation Study}
\label{section4.4}
\textbf{Effectiveness of the variance consistency loss.} We compare the variance consistency loss ${{\mathcal{L}}_{v}}$ with the distance loss ${{\mathcal{L}}_{d}}$ (in Equation.~\ref{ex5}) to verify the effectiveness of ${{\mathcal{L}}_{v}}$. For each loss, we choose a set of suitable weights $\lambda$ separately to represent its trade-off between the protection quality and recovery quality. Its results are shown in Figure.~\ref{fig7}. The protected image generated using ${{\mathcal{L}}_{d}}$ has a smaller SSIM value compared to the protected image using ${{\mathcal{L}}_{v}}$, but a clear outline can be seen. So the visual information of the original image still exists. If the protected image is to be made free of such visual information, there is a significant loss in the quality of the recovered image. In contrast, using ${{\mathcal{L}}_{v}}$, it is possible to generate protected images with both no residual visual information and high recovery quality. We also use another key to recover the protected image to further investigate the effect of the two different losses on the protection quality. From Figure.~\ref{fig7}, it can be concluded that the protected image using ${{\mathcal{L}}_{v}}$ is more difficult to be recovered by similar key models with different initialization while ensuring the quality of the recovered image. We compare $\mathcal{L}_v$ with more other losses in Supplementary Material D to further demonstrate the effectiveness of $\mathcal{L}_v$.

As shown in Figure.~\ref{fig5}, the difference between the results when $\lambda=0.001$ and $\lambda=0.01$ is very slight for the protected and recovery quality of the protected images. This means that our method requires no carefully adjustment of the weight $\lambda$ for ${{\mathcal{L}}_{v}}$, which is very convenient.

\begin{figure}[t]
  \centering
  \includegraphics[width=0.99\columnwidth]{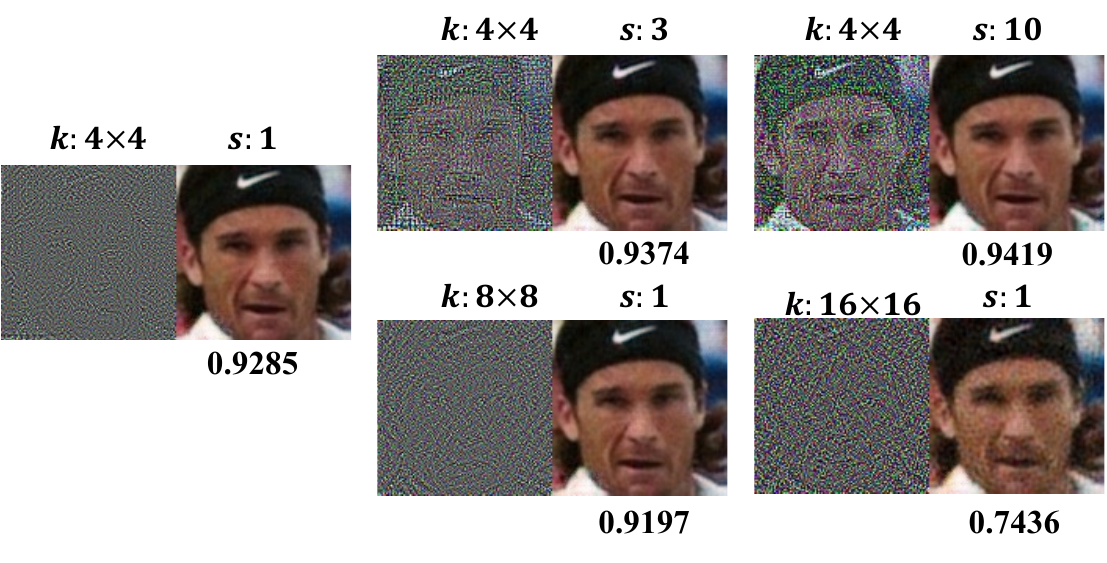}
  \caption{Impact of parameter settings on protection quality and recovery quality. $k$ represents the size of the convolution kernel and $s$ represents the step size. For each pair of images, the former is the protected image and the latter is the recovery image. Below the recovery image is marked the SSIM value between it and the original image.}
  \label{fig20}
  %\vskip -0.1in
\end{figure}

\textbf{Parameters of variance consistency loss.} In the variance consistency loss calculation, the process of blocking the image and calculating the internal sum of each block can be regarded as a convolution kernel with all values of 1 to convolve the image. The convolution kernel size represents the size of the block, and the step size of the convolution kernel represents the degree of overlap of the block. We explored the effect of block size and degree of overlap on performance, as shown in Figure~\ref{fig20}. When the overlap degree of blocks is certain (especially when it is highly overlapping), the quality of the recovery image decreases as the size of the blocks increases. When the size of blocks is certain, the protection quality decreases as the overlap degree of blocks decreases. In this work, we divide the image into $4\times4$ blocks, with 12 pixels overlapping between adjacent blocks (step size is 1). This setting is applicable to most tasks and there is no need to change it for different samples.

%-------------------------------------------------------------------------
\begin{figure}[t]
  \begin{center}
  \includegraphics[width=0.99\columnwidth]{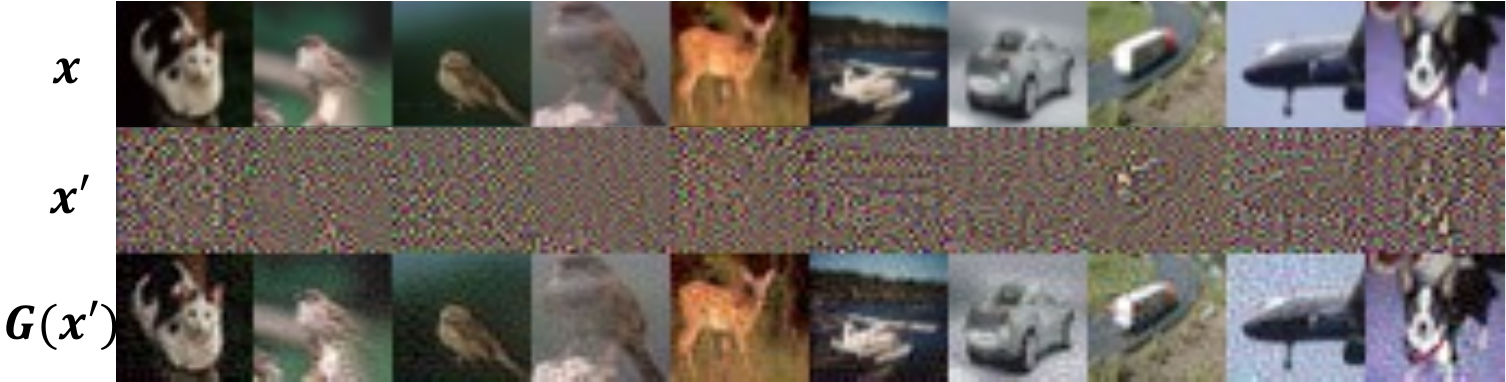}
  \end{center}
  \caption{Results of AVIH method for Resnet50 as the service model. Each column represents a different sample.}
  \label{fig8}
  \vskip -0.01in
\end{figure}

\subsection{Privacy protection for classification models}

We trained ResNet50 \cite{ref47}, VGG19 \cite{ref48} using the train set of \textit{CIFAR-10} \cite{ref43} and implemented AVIH method for the images of the test set. The results are shown in Figure.~\ref{fig8}. We also compared two existing methods suitable for visual information hiding and the results are shown in Table.~\ref{tab:4}, where the ITP \cite{ref10} do not recover protected images. Since the LIE \cite{ref6} uses protected images to train the model, the prediction accuracy on the original data is low. We show the visual metrics, the results of the two comparative methods, and the results on Imgnet in Supplementary Material C. Our method has minimal impact on the accuracy of the models, and the average SSIM value of the recovered images can reach above 0.9 for both models after our tests.

\begin{table}[t]
  \caption{Impact of privacy protection methods on the accuracy (percentage) of classification models on \textit{CIFAR-10}.}
  \label{tab:4}
  \renewcommand\tabcolsep{8.0pt}
  \renewcommand\arraystretch{1.0}
\begin{center}
\begin{small}
  \begin{tabular}{l|l|cccc}
  \hline
  Method&Model& Original ($\%$) &Protect ($\%$)\\
  \hline
  \multirow{2}{*}{LIE \cite{ref6}}&VGG19 & 10.59&87.78\\
  ~ &Resnet50 & 11.00&91.53\\
  \hline
  \multirow{2}{*}{ITP \cite{ref10}}&VGG19 & 93.95&90.70\\
  ~ &Resnet50 & 95.53&90.16\\
  \hline
  \multirow{2}{*}{AVIH (Our)}&\cellcolor{lightgray!40}VGG19 &\cellcolor{lightgray!40}93.95&\cellcolor{lightgray!40}\textbf{93.95}\\
  ~ &\cellcolor{lightgray!40}Resnet50 &\cellcolor{lightgray!40}95.53&\cellcolor{lightgray!40}\textbf{95.28}\\
  \hline
  \end{tabular}
  \end{small}
  \end{center}
  %\vskip -0.15in
\end{table}

\section{Conclusion}
In this paper, we propose a visual information hiding method AVIH based on Type-I attack. We evaluate our method by image protection of face recognition systems in the cloud. Experiments show that the AVIH method can protect images while preserving their functionality for the service model. We also propose a variance consistency loss to solve the problem of the difficult trade-off between protection quality and recovery quality. Finally, we use the AVIH method in classification tasks with satisfactory results.
Our work provides a new perspective on image visual information protection, which has beneficial implications for adversarial learning communities and private protection communities. The AVIH method requires complete service model information. While this is feasible in most image storage situations, it limits the applicability of the AVIH method to a wider range of image protection scenarios. This is also a problem we will solve in future work.

\vspace{1em}
\noindent \textbf{Acknowledgement.} \quad 
This work was supported in part by the National Natural Science Foundation of China under Grants U22A2096 and 62036007; in part by the Fundamental Research Funds for the Central Universities under Grants QTZX23042 and XJS221502; in part by the Technology Innovation Leading Program of Shaanxi under Grant 2022QFY01-15; in part by Open Research Projects of Zhejiang Lab under Grant 2021KG0AB01; in part by the Natural Science Basis Research Plan in Shaanxi Province of China under Grant 2022JQ-696; in part by the Fundamental Research Funds for the Central Universities and in part by the Innovation Fund of Xidian University under Grant YJSJ23012.
%-------------------------------------------------------------------------

%%%%%%%%% REFERENCES
{\small
\bibliographystyle{ieee_fullname}
\bibliography{egbib}
}
%-------------------------------------------------------------------------

\appendix

\section*{Supplementary Material} 

\section{Time Spent on Protection}
\begin{figure*}[t]
  \centering
  \includegraphics[width=0.7\textwidth]{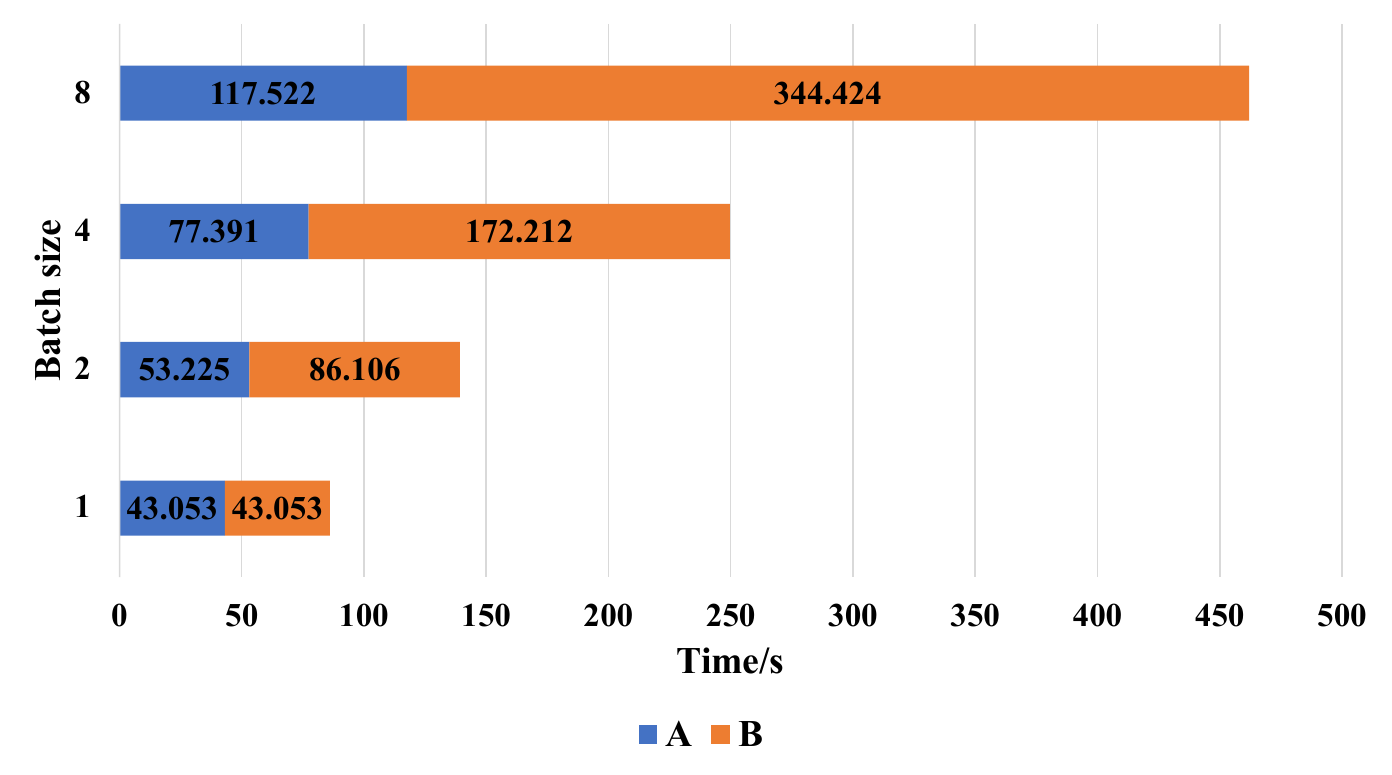}
  \caption{The time required to protect an image by the AVIH method and the effect of batchsize on protection time. A represents the average time required to protect a batch, and B represents the average time required to protect the corresponding number of samples when the batch size is 1. We protected ten batches and averaged the time spent.}
  \label{fig18}
  %\vskip -0.1in
\end{figure*}

\begin{figure*}[t]
  \centering
  \includegraphics[width=0.85\textwidth]{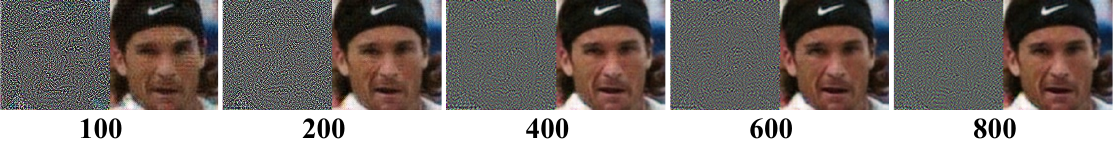}
  \caption{Protected and recovery images generated by different iterations. Each pair of images includes the protected image and recovery image, respectively. The number of iterations they correspond to is marked below the image.}
  \label{fig19}
  %\vskip -0.1in
\end{figure*}

The GPU model used for all our experiments is NVIDIA TITAN Xp. Since our method is an iterative method, it takes a certain amount of time to protect images one by one. However, we can increase the protection efficiency by increasing the batch size. As shown in Figure.~\ref{fig18}, we tested the average time it takes to protect an image under different batch sizes, from which we can conclude that adjusting the batch size can greatly improve the efficiency of protection. Moreover, the number of iterations also greatly affects the time required for protection. We tested the effect of different iterations on the time required for protection and the quality of protected images as shown in Table~\ref{tab:12} and Figure.~\ref{fig19}. We can conclude that, within a certain range, we can speed up the protection time with little impact on protection quality by reducing the number of iterations. We can also spend more time increasing the protection quality by increasing the number of iterations.

\section{Security Analysis}
\label{Appen1}
\textbf{Key Model Space Analysis.} In our proposed method, we choose the generative model as the key, which has a large key space. Take the key model used in our experiments as an example, it has a size of 11.383M. Such a large key space can be disastrous for those who use exhaustive attacks.

\textbf{Histogram Analysis and Correlation Analysis.} Plain images have a strong correlation between two adjacent pixels in the horizontal and vertical directions\cite{ref49}, and protection methods with good properties often need to break this correlation\cite{ref50}. Therefore, we performed a correlation analysis of our proposed method, and its results are shown in Figure.~\ref{fig20}. Compared to the strong correlation of the original image, the protected image we obtained greatly reduces the correlation between adjacent pixels. We also performed histogram analysis on the original and protected images, and the results are shown in Figure.~\ref{fig21}. From it, it can be concluded that the histogram statistical properties of the protected image and the original image are completely different, and the protected image more closely resembles a gaussian distribution. So the protected images obtained by the AVIH method have well statistical characteristics.

%-----------------

\textbf{Identifiability of Protected Images.} We use one face recognition model as the target model to generate protected images, and then use another face recognition model to identify the results obtained in Table~\ref{tab:11}. It can be concluded that the protected image obtained by a target model cannot be used normally by other face recognition models. Storing such protected images in a cloud environment can greatly improve security.

\section{Privacy Protection for Classification Tasks}
\label{Appen2}
\textbf{Detailed Experimental Setup.} When implementing the AVIH method on the classification task, we adjust the batch size to 10 and initialize the protected image as the original image. We set the number of iterations to 600.

\textbf{The Quality of Recovered Images.} We show the protection results of LIE \cite{ref6} and ITP \cite{ref10} in Figure.~\ref{fig15-16}. We tested the average SSIM values of the AVIH method for the protected and recovery images of the test set in \textit{CIFAR-10} \cite{ref43}. The results are shown in Table.~\ref{tab:10}.  We compare our results with LIE \cite{ref6} and ITP \cite{ref10}, where LIE recovers the original image, but the protection quality is weaker and has a significant impact on the model accuracy. ITP mitigates the impact of the protected image on the model accuracy, but the protected image becomes unrecoverable. And our method ensures a strong protection strength while the impact on the model accuracy is very slight.

\textbf{Test on IamgeNet.} We set the target model as ResNet50 and selected 50 images in ImageNet for testing as in Table~\ref{tab:b}.

\section{Details of Variance Consistency Loss}
\textbf{Motivation.} We replaced variance consistency loss with different losses and test the impact of these losses on visual information hiding separately. The results are shown in Figure.~\ref{fig22}, where VC Loss represents our proposed variance consistency loss, MSE Loss represents the mean square error loss of the protected image and the original image, T Loss represents the mean square error of the protected image and the full gray image, and TV Loss represents the total variation loss of the protected image.
From it, it can be concluded that the conventional loss focus on the difference between pixels. Images obtained by maximizing MSE between the protected image and the original image still retain some spatial features of the original image (\eg, the contours of a human face) despite large pixel changes (\eg, color textures), as shown in Figure.~\ref{fig22}. To solve this issue, we consider making the pixel distribution of the protected image as consistent as possible at each location, so that the protected image cannot exhibit obvious spatial features in the pixel space. However, it is difficult (and unnecessary) to make every pixel converge to the same value. We thus block the image so that the pixel distribution between each block is similar while giving more possibilities for variation of pixels within the block. 

% \textbf{Parameters.} In the variance consistency loss calculation, the process of blocking the image and calculating the internal sum of each block can be regarded as a convolution kernel with all values of 1 to convolve the image. The convolution kernel size represents the size of the block, and the step size of the convolution kernel represents the degree of overlap of the block. We explored the effect of block size and degree of overlap on performance, as shown in Figure~\ref{fig20}. When the overlap degree of blocks is certain (especially when it is highly overlapping), the quality of the recovery image decreases as the size of the blocks increases. When the size of blocks is certain, the protection quality decreases as the overlap degree of blocks decreases. In this work, we divide the image into $4\times4$ blocks, with 8 pixels overlapping between adjacent blocks (step size is 1). This setting is applicable to most tasks.
%-------------------------------------------

%----------------------------------------------------------------
\begin{table*}[t!]
  \renewcommand\tabcolsep{12pt}
  \renewcommand\arraystretch{1.2}
  \begin{center}
  %\begin{small}
  \caption{The effect of different iterations on protection time and protected image quality. Time represents the average time required to protect a batch, SSIM represents the average SSIM between the recovery image and the original image, and COS represents the average cosine similarity between the original image feature and the protected image features. We tested 10 samples and averaged these metrics.}
  \label{tab:12}
  \begin{tabular}{c|ccccc}
  \toprule 
  Iterations& 100& 200& 400& 600& 800\\
  \midrule
  Time(s)& 5.651& 11.251& 22.703& 32.703& 43.053\\
  SSIM& 0.721& 0.829& 0.886& 0.895& 0.912\\
  COS& 0.973& 0.983& 0.995& 0.997& 0.997\\
  \bottomrule 
  \end{tabular}
  %\end{small}
  \end{center}
%\vskip -0.1in
\end{table*}

\begin{figure*}[!t]
\centering

\begin{tikzpicture}
\matrix[column sep=0.01cm, row sep=0.01cm]{
  \node {\includegraphics[width=0.48\textwidth]{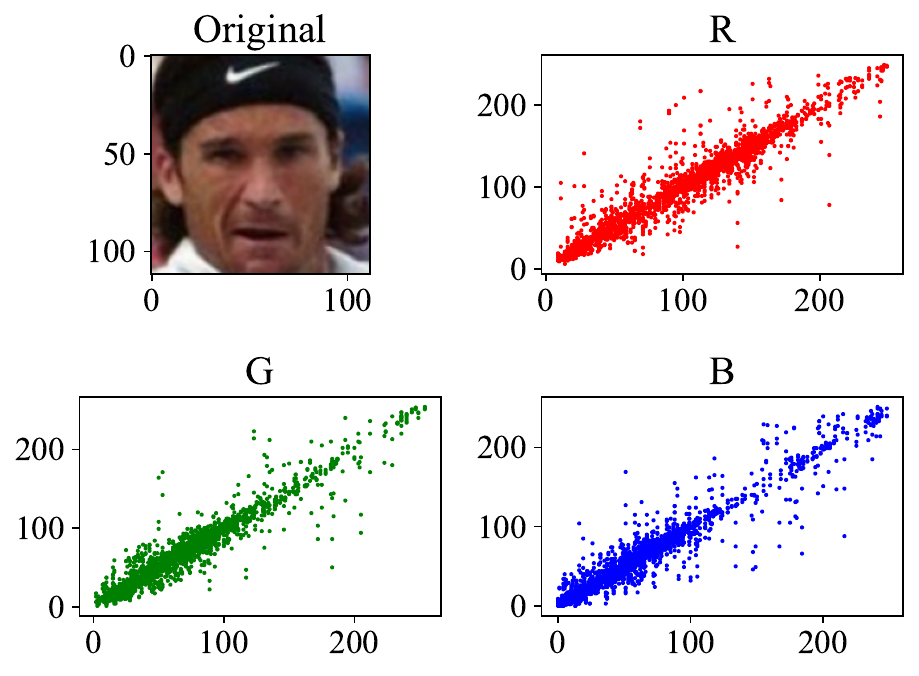}}; &
  \node {\includegraphics[width=0.48\textwidth]{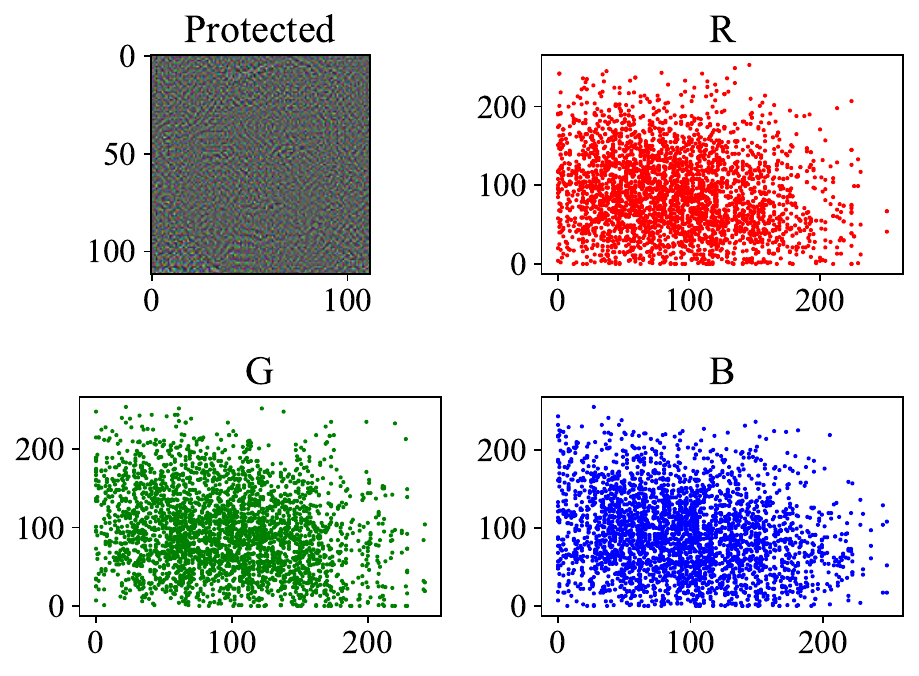}}; \\
};

\end{tikzpicture}
\caption{Results of correlation analysis. We randomly select 1000 pairs of adjacent pixel points for the original and protected images, respectively, and calculate their correlation coefficients in horizontal, vertical and diagonal directions.}
\label{fig20}
\end{figure*}

\begin{figure*}[!t]
\centering
\begin{tikzpicture}
\matrix[column sep=0.01cm, row sep=0.01cm]{
  \node {\includegraphics[width=0.48\textwidth]{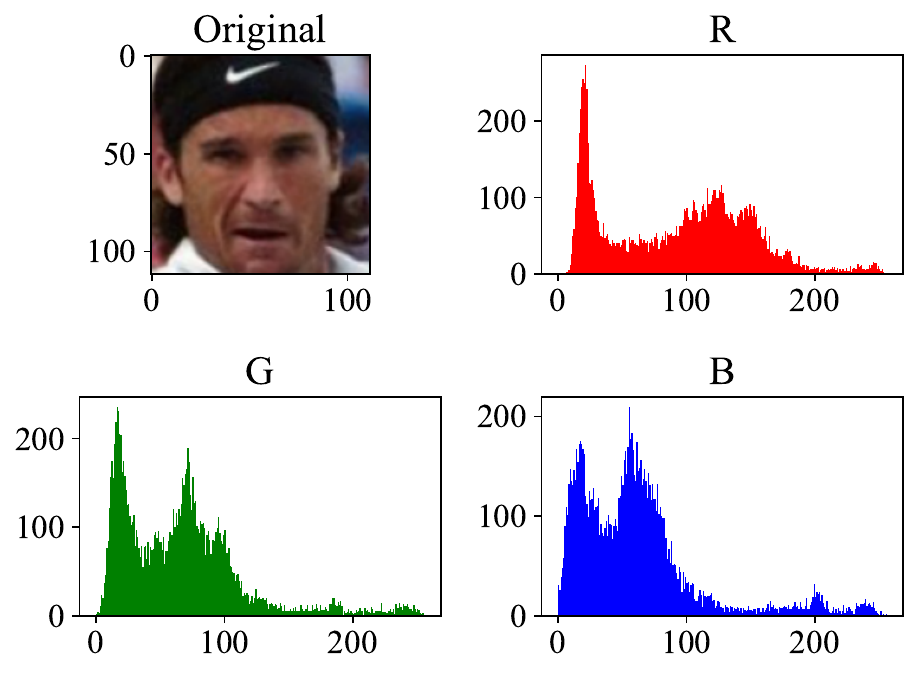}}; &
  \node {\includegraphics[width=0.48\textwidth]{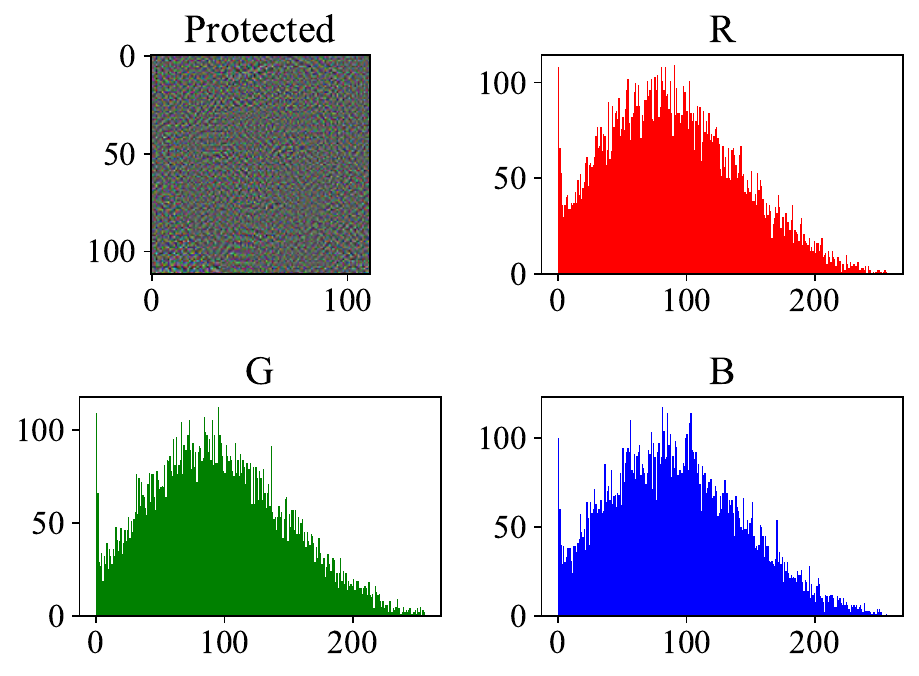}}; \\
  \node {\includegraphics[width=0.48\textwidth]{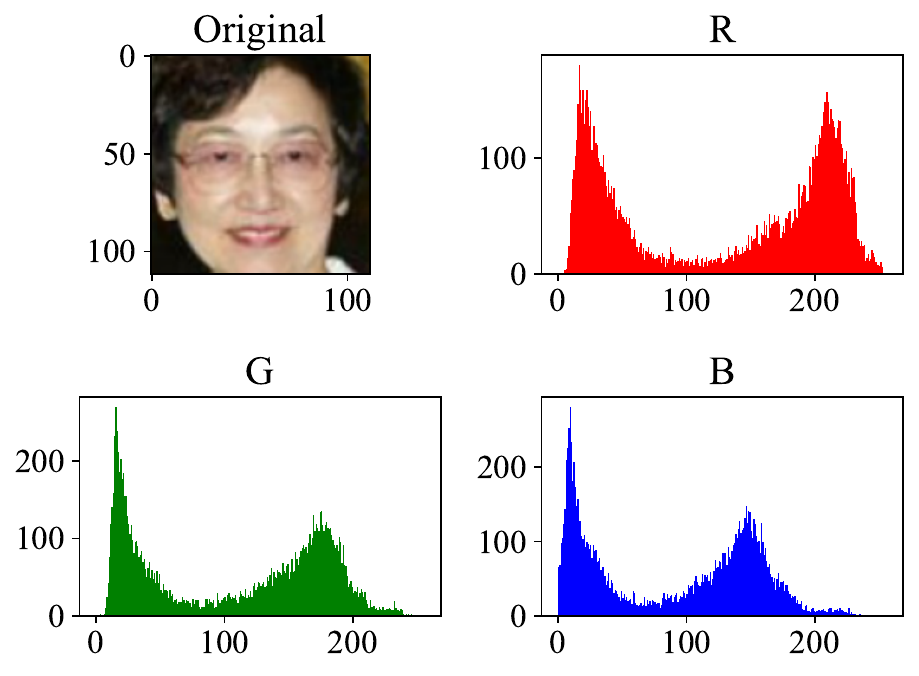}}; &
  \node {\includegraphics[width=0.48\textwidth]{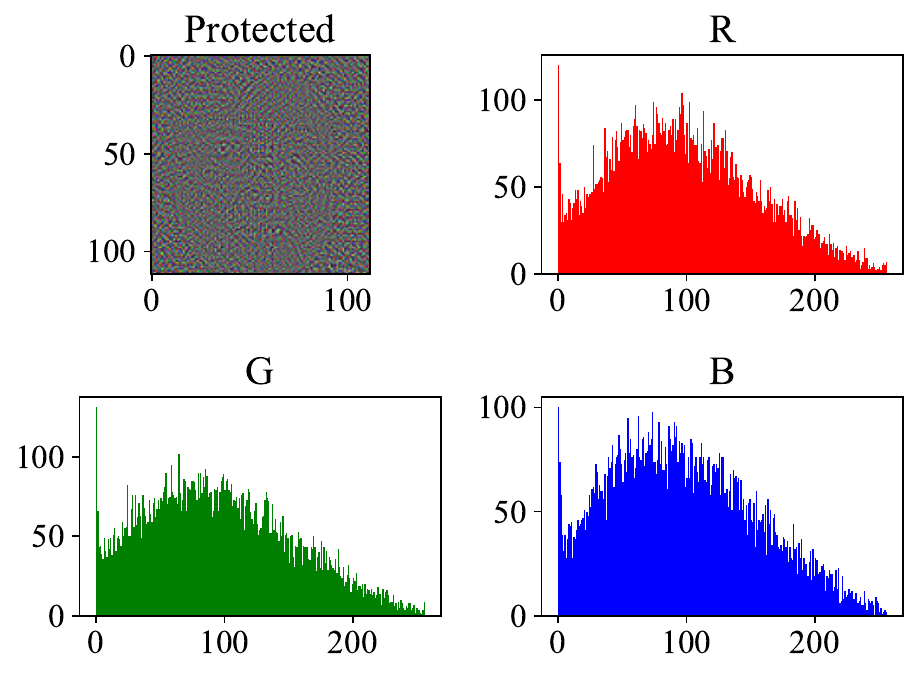}}; \\
};
\end{tikzpicture}
\caption{Results of histogram analysis of some original and protected images. The original image is on the left and its corresponding protected image is on the right.}
\label{fig21}
\end{figure*}

% \begin{figure*}[!t]
% \centering
% \begin{subfigure}{0.45\linewidth}
% \includegraphics[width=.9\linewidth]{fig9.pdf}%
% \label{fig_first_case}
% \end{subfigure}
% \hfil
% \begin{subfigure}{0.45\linewidth}
% \includegraphics[width=.9\linewidth]{fig10.pdf}%
% \label{fig_second_case}
% \end{subfigure}
% \caption{Results of correlation analysis. We randomly select 1000 pairs of adjacent pixel points for the original and protected images, respectively, and calculate their correlation coefficients in horizontal, vertical and diagonal directions.}
% \label{fig20}
% \end{figure*}

% \begin{figure*}[!t]
% \centering
% \begin{subfigure}{0.48\linewidth}
% \includegraphics[width=.9\linewidth]{fig11.pdf}%
% \label{fig_first_case_1}
% \end{subfigure}
% \hfil
% \begin{subfigure}{0.48\linewidth}
% \includegraphics[width=.9\linewidth]{fig12.pdf}%
% \label{fig_second_case_2}
% \end{subfigure}
% \hfil  
% \begin{subfigure}{0.48\linewidth}
% \includegraphics[width=.9\linewidth]{fig13.pdf}%
% \label{fig_second_case_3}
% \end{subfigure}
% \hfil
% \begin{subfigure}{0.48\linewidth}
% \includegraphics[width=.9\linewidth]{fig14.pdf}%
% \label{fig_second_case_4}
% \end{subfigure}
% \caption{Results of histogram analysis of some original and protected images. The original image is on the left and its corresponding protected image is on the right.}
% \label{fig21}
% \end{figure*}

\begin{figure*}[t]
  \centering
  \includegraphics[width=0.9\textwidth]{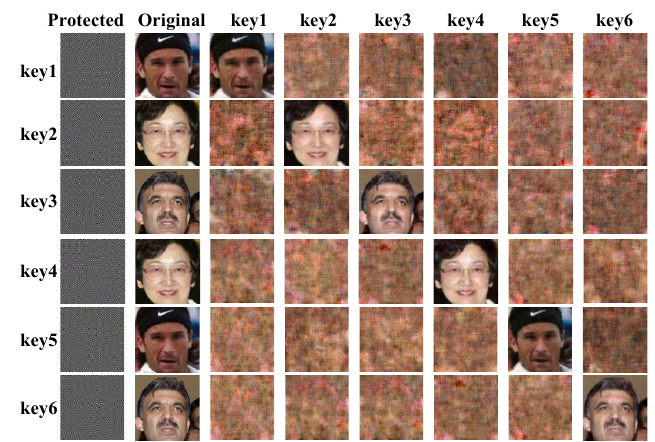}
  \caption{Results of the key model randomness analysis. Due to space limitations, we show the results for 6 key models.}
  \label{fig17}
  \vskip -0.1in
\end{figure*}

% \begin{figure*}[t]
%   \centering
% \begin{subfigure}{0.9\linewidth}
% \includegraphics[width=.9\linewidth]{fig15.pdf}%
% \label{fig15}
% \end{subfigure}
% \begin{subfigure}{0.9\linewidth}
% \includegraphics[width=.9\linewidth]{fig16.pdf}%
% \label{fig16}
% \end{subfigure} 
%  \caption{Results of the LIE method and the ITP method. The top group is the result of the LIE method, and the bottom group is the result of the ITP method. Each column represents a different sample.}
% \label{fig15-16}
% \end{figure*}

\begin{figure*}[t]
  \centering
  \begin{tikzpicture}
    \matrix[column sep=0.01cm, row sep=0.01cm]{
      \node {\includegraphics[width=0.78\textwidth]{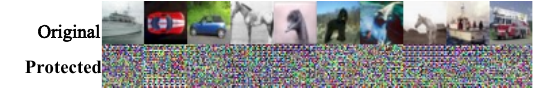}}; \\
      \node {\includegraphics[width=0.78\textwidth]{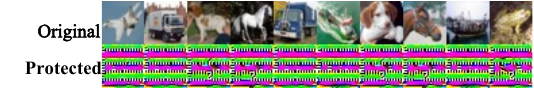}}; \\
    };
  \end{tikzpicture}
 \caption{Results of the LIE method and the ITP method. The top group is the result of the LIE method, and the bottom group is the result of the ITP method. Each column represents a different sample.}
\label{fig15-16}
\end{figure*}

\begin{table*}[t]
  \renewcommand\tabcolsep{12pt}
  \renewcommand\arraystretch{1.2}
  \begin{center}
  %\begin{small}
  \caption{Accuracy (percentage) of predicting protected images using a model different from the target face recognition model. The same name represents the same model.}
  \label{tab:11}
  \begin{tabular}{c|cccc}
  \toprule 
  \diagbox{Target model}{Test model}&AdaFace&ArcFace&CosFace&SphereFce\\
  \midrule
  AdaFace&98.6 & 0&-&-\\
  ArcFace&0 &96.5&-&-\\
  \midrule
  CosFace&- & -&89.4&0\\
  SphereFce&- &-& 0&80.3\\
  \bottomrule 
  \end{tabular}
  %\end{small}
  \end{center}
\vskip -0.1in
\end{table*}

\begin{table*}[t!]
  \renewcommand\tabcolsep{12pt}
  \renewcommand\arraystretch{1.2}
  \begin{center}
  %\begin{small}
  \caption{Image visual quality metrics of visual information hiding methods of classification models on \textit{CIFAR-10}.}
  \label{tab:10}
  \begin{tabular}{c|c|cccc}
  \toprule 
  Method&Model& SSIM$_{e}$ &SSIM$_{d}$\\
  \midrule
  \multirow{2}{*}{LIE \cite{ref6}}&VGG19 & 0.178&1.000\\
  ~ &Resnet50 & 0.178&1.000\\
  \midrule
  \multirow{2}{*}{ITP \cite{ref10}}&VGG19 & 0.068&-\\
  ~ &Resnet50 & 0.073&-\\
  \midrule
  \multirow{2}{*}{AVIH(our)}&\cellcolor{lightgray!40}VGG19 &\cellcolor{lightgray!40}0.171&\cellcolor{lightgray!40}0.900\\
  ~ &\cellcolor{lightgray!40}Resnet50 &\cellcolor{lightgray!40}0.198&\cellcolor{lightgray!40}0.923\\
  \bottomrule 
  \end{tabular}
  %\end{small}
  \end{center}
\vskip -0.1in
\end{table*}

\begin{table*}[t]
\renewcommand\tabcolsep{12pt}
\renewcommand\arraystretch{1.2}
\begin{center}
%\begin{footnotesize}
\caption{Evaluation results of 50 samples in ImageNet. SSIM$_{d}$ represents the SSIM between the original and the recovered image.}
\label{tab:b}
\begin{tabular}{l|ccc}
\toprule 
Model & Original Acc.(\%) & Protect Acc.(\%) & SSIM$_{d}$\\
\midrule
ResNet50 & 0.86 & 0.86 & 0.90 \\
\bottomrule
\end{tabular}
%\end{footnotesize}
\end{center}
%\vskip -0.3in
\end{table*}

\begin{figure*}[t]
  \centering
  \includegraphics[width=0.9\textwidth]{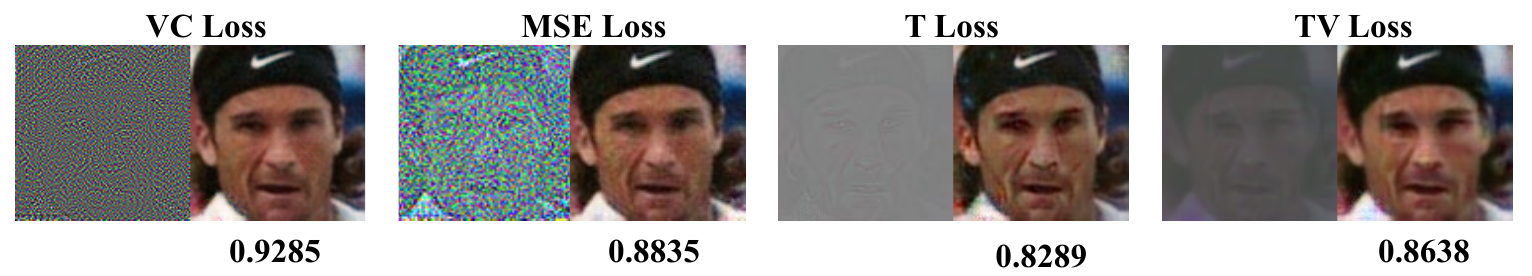}
  \caption{Impact of different losses on protected quality and recovery quality. For each pair of images, the former is the protected image and the latter is the recovery image. Below the recovery image is marked the SSIM value between it and the original image.}
  \label{fig22}
  \vskip -0.1in
\end{figure*}

\end{document}